\documentclass[10pt,journal,compsoc]{IEEEtran}
\usepackage[table,dvipsnames]{xcolor}
\usepackage{tikz}
\ifCLASSOPTIONcompsoc
  \usepackage[nocompress]{cite}
\else
  \usepackage{cite}
\fi
\ifCLASSINFOpdf
\else
\fi
\usepackage{pgfplots}
\pgfplotsset{compat=newest}
\usepackage{booktabs} % Nice tables
\usepackage{etoolbox,siunitx} % Align decimal point
\usepackage{pgfplotstable} % For plots
\usetikzlibrary{plotmarks} % For plots
\usetikzlibrary{colorbrewer} % Nice colors
\usepackage{adjustbox}

\usepackage{rotating}
\usepackage{graphicx}
\usepackage{amsmath,amssymb} % define this before the line numbering.
\usepackage{url}

\usepackage{color}
\usepackage{pifont}
\definecolor{olivegreen}{RGB}{0,170,0}

\newcommand{\J}{\mathcal{J}}
\newcommand{\F}{\mathcal{F}}

\definecolor{rowblue}{RGB}{220,230,240}

\usepackage{subfiles}

\usepackage{booktabs}
\usepackage{multicol}
\definecolor{rowblue}{RGB}{220,230,240}

\usepackage{capt-of}

\begin{document}
%
% paper title
% Titles are generally capitalized except for words such as a, an, and, as,
% at, but, by, for, in, nor, of, on, or, the, to and up, which are usually
% not capitalized unless they are the first or last word of the title.
% Linebreaks \\ can be used within to get better formatting as desired.
% Do not put math or special symbols in the title.
\title{The 2018 DAVIS Challenge on\\Video Object Segmentation}
%
%
% author names and IEEE memberships
% note positions of commas and nonbreaking spaces ( ~ ) LaTeX will not break
% a structure at a ~ so this keeps an author's name from being broken across
% two lines.
% use \thanks{} to gain access to the first footnote area
% a separate \thanks must be used for each paragraph as LaTeX2e's \thanks
% was not built to handle multiple paragraphs
%
%
%\IEEEcompsocitemizethanks is a special \thanks that produces the bulleted
% lists the Computer Society journals use for "first footnote" author
% affiliations. Use \IEEEcompsocthanksitem which works much like \item
% for each affiliation group. When not in compsoc mode,
% \IEEEcompsocitemizethanks becomes like \thanks and
% \IEEEcompsocthanksitem becomes a line break with idention. This
% facilitates dual compilation, although admittedly the differences in the
% desired content of \author between the different types of papers makes a
% one-size-fits-all approach a daunting prospect. For instance, compsoc
% journal papers have the author affiliations above the "Manuscript
% received ..."  text while in non-compsoc journals this is reversed. Sigh.

\author{Sergi Caelles, Alberto Montes, Kevis-Kokitsi Maninis, Yuhua Chen,\\
        Luc Van Gool, Federico Perazzi, and~Jordi Pont-Tuset% <-this % stops a space
\IEEEcompsocitemizethanks{\IEEEcompsocthanksitem S. Caelles, A. Montes, K.-K. Maninis, Y. Chen, L. Van Gool, and J. Pont-Tuset are with the Computer Vision Laboratory, ETH Z\"urich, Switzerland.\protect\\
\IEEEcompsocthanksitem F. Perazzi is with Disney Research, Z\"urich, Switzerland.\protect\\
% note need leading \protect in front of \\ to get a newline within \thanks as
% \\ is fragile and will error, could use \hfil\break instead.
}% <-this % stops an unwanted space
\thanks{Contacts and updated information can be found in the challenge website: http://davischallenge.org}}

% note the % following the last \IEEEmembership and also \thanks -
% these prevent an unwanted space from occurring between the last author name
% and the end of the author line. i.e., if you had this:
%
% \author{....lastname \thanks{...} \thanks{...} }
%                     ^------------^------------^----Do not want these spaces!
%
% a space would be appended to the last name and could cause every name on that
% line to be shifted left slightly. This is one of those "LaTeX things". For
% instance, "\textbf{A} \textbf{B}" will typeset as "A B" not "AB". To get
% "AB" then you have to do: "\textbf{A}\textbf{B}"
% \thanks is no different in this regard, so shield the last } of each \thanks
% that ends a line with a % and do not let a space in before the next \thanks.
% Spaces after \IEEEmembership other than the last one are OK (and needed) as
% you are supposed to have spaces between the names. For what it is worth,
% this is a minor point as most people would not even notice if the said evil
% space somehow managed to creep in.

% The paper headers
\markboth{}%
{}
% The only time the second header will appear is for the odd numbered pages
% after the title page when using the twoside option.
%
% *** Note that you probably will NOT want to include the author's ***
% *** name in the headers of peer review papers.                   ***
% You can use \ifCLASSOPTIONpeerreview for conditional compilation here if
% you desire.

% The publisher's ID mark at the bottom of the page is less important with
% Computer Society journal papers as those publications place the marks
% outside of the main text columns and, therefore, unlike regular IEEE
% journals, the available text space is not reduced by their presence.
% If you want to put a publisher's ID mark on the page you can do it like
% this:
%\IEEEpubid{0000--0000/00\$00.00~\copyright~2015 IEEE}
% or like this to get the Computer Society new two part style.
%\IEEEpubid{\makebox[\columnwidth]{\hfill 0000--0000/00/\$00.00~\copyright~2015 IEEE}%
%\hspace{\columnsep}\makebox[\columnwidth]{Published by the IEEE Computer Society\hfill}}
% Remember, if you use this you must call \IEEEpubidadjcol in the second
% column for its text to clear the IEEEpubid mark (Computer Society jorunal
% papers don't need this extra clearance.)

% use for special paper notices
%\IEEEspecialpapernotice{(Invited Paper)}

% for Computer Society papers, we must declare the abstract and index terms
% PRIOR to the title within the \IEEEtitleabstractindextext IEEEtran
% command as these need to go into the title area created by \maketitle.
% As a general rule, do not put math, special symbols or citations
% in the abstract or keywords.
\IEEEtitleabstractindextext{%
\begin{abstract}
We present the \emph{2018 DAVIS Challenge on Video Object Segmentation}, a public competition 
specifically designed for the task of video object segmentation.
It builds upon the DAVIS 2017 dataset, which was presented in the previous edition of the DAVIS Challenge~\cite{Pont-Tuset_arXiv_2017}, and added 100 videos with multiple objects per sequence to the original DAVIS 2016 dataset~\cite{Perazzi2016}.
Motivated by the analysis of the results of the 2017 edition~\cite{Pont-Tuset_arXiv_2017}, the main track of the competition will be the same than in the previous edition (segmentation given the full mask of the objects in the first frame -- semi-supervised scenario). 
This edition, however, also adds an interactive segmentation teaser track, where the participants will interact with a web service simulating the input of a human that provides scribbles to iteratively improve the result.
\end{abstract}

% Note that keywords are not normally used for peerreview papers.
}

% make the title area
\maketitle

% To allow for easy dual compilation without having to reenter the
% abstract/keywords data, the \IEEEtitleabstractindextext text will
% not be used in maketitle, but will appear (i.e., to be "transported")
% here as \IEEEdisplaynontitleabstractindextext when the compsoc
% or transmag modes are not selected <OR> if conference mode is selected
% - because all conference papers position the abstract like regular
% papers do.
\IEEEdisplaynontitleabstractindextext
% \IEEEdisplaynontitleabstractindextext has no effect when using
% compsoc or transmag under a non-conference mode.

% For peer review papers, you can put extra information on the cover
% page as needed:
% \ifCLASSOPTIONpeerreview
% \begin{center} \bfseries EDICS Category: 3-BBND \end{center}
% \fi
%
% For peerreview papers, this IEEEtran command inserts a page break and
% creates the second title. It will be ignored for other modes.
\IEEEpeerreviewmaketitle

\section{Introduction}
The Densely-Annotated VIdeo Segmentation (DAVIS) initiative~\cite{Perazzi2016} supposed a 
significant increase in the size and quality of the benchmarks for video object
segmentation at the time. 
The availability of such a dataset was key in the appearance of new techniques in the 
field~\cite{voigtlaender17BMVC,Caelles2017,Perazzi2017,Cheng2017,Jang2017,Jampani2017,Koh2017,Tokmakov2017,Jain2017,Tokmakov2017a}
that boosted the performance of the state of the art.
The 2017 DAVIS Challenge on Video Object Segmentation~\cite{Pont-Tuset_arXiv_2017} presented an extension of the dataset: up to 150 sequences (10474 annotated frames) from 50 sequences (3455 frames), more than one annotated object per sequence (384 objects instead of 50), and more challenging scenarios such as motion, occlusions, etc.
This initiative again supposed a boost in 20\% better results in the state of the art (see the analysis in~\cite{Pont-Tuset_arXiv_2017}).

Motivated by the success of the first edition of the challenge, this paper presents the 2018 DAVIS Challenge on Video Object Segmentation, whose results will be presented in a workshop co-located with CVPR 2018, in Salt Lake City, USA. 
The main track will be, as in the previous edition, the semi-supervised segmentation, where the mask of the object in the first frame is given to the algorithm, and the result is the segmentation in the rest of the frames. The dataset and its partitions will also be kept as in the 2017 edition.

Apart from the main track, we introduce a new \textit{teaser} track, in which we will explore the evaluation of \textbf{interactive video object segmentation} algorithms, that is, a scenario where the user provides a very simple and quick input to the algorithm, waits for the first result; then provides another input in order to refine the current result and iterates until the result is satisfactory.
The main motivation behind this new scenario is twofold.
First, the semi- and unsupervised scenarios represents two extremes of the level of user interaction with the algorithm: the former needs a pixel-level accurate segmentation of the first frame (very time consuming for a human to provide)
and the latter does not take any user input into account.
Second, the benchmark does not take speed into account, that is, how much the user must wait until the result is obtained after marking the first segmentation.

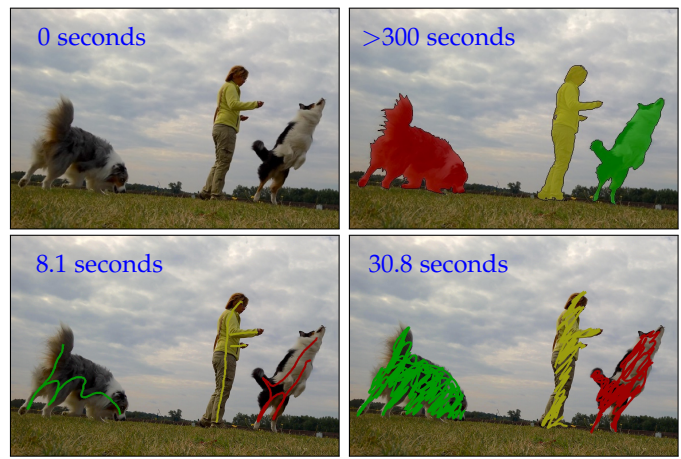
\begin{figure}[t]
\label{fig:levels_of_interaction}
\setlength{\fboxsep}{0pt}
\begin{tikzpicture}
    \draw (0, 0) node[inner sep=0] {\fbox{\adjincludegraphics[width=0.49\linewidth,trim={{.03\width} {.1\height} {.3\width} {.1\height}},clip]{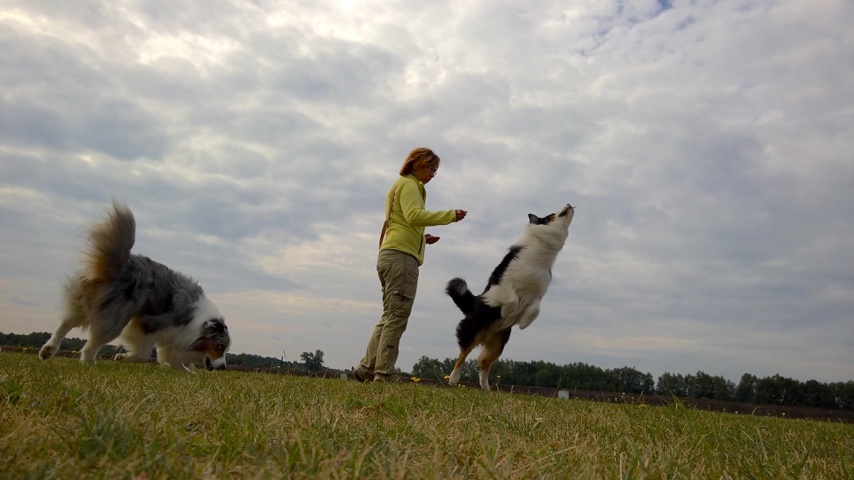}}};
    \draw (-1.1, 1.1) node[color=blue] {0 seconds};
\end{tikzpicture}
\hfill
\begin{tikzpicture}
    \draw (0, 0) node[inner sep=0] {\fbox{\adjincludegraphics[width=0.49\linewidth,trim={{.03\width} {.1\height} {.3\width} {.1\height}},clip]{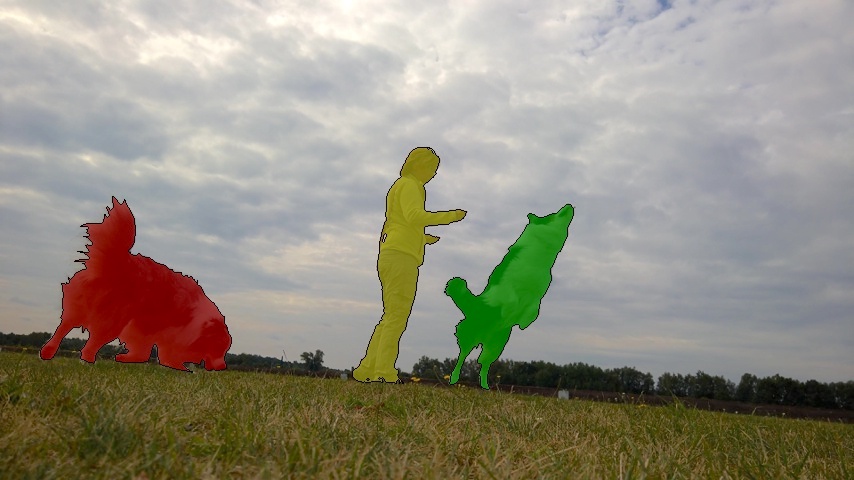}}};
    \draw (-1, 1.1) node[color=blue] {$>$300 seconds};
\end{tikzpicture}\\[2pt]
\begin{tikzpicture}
    \draw (0, 0) node[inner sep=0] {\fbox{\adjincludegraphics[width=0.49\linewidth,trim={{.03\width} {.1\height} {.3\width} {.1\height}},clip]{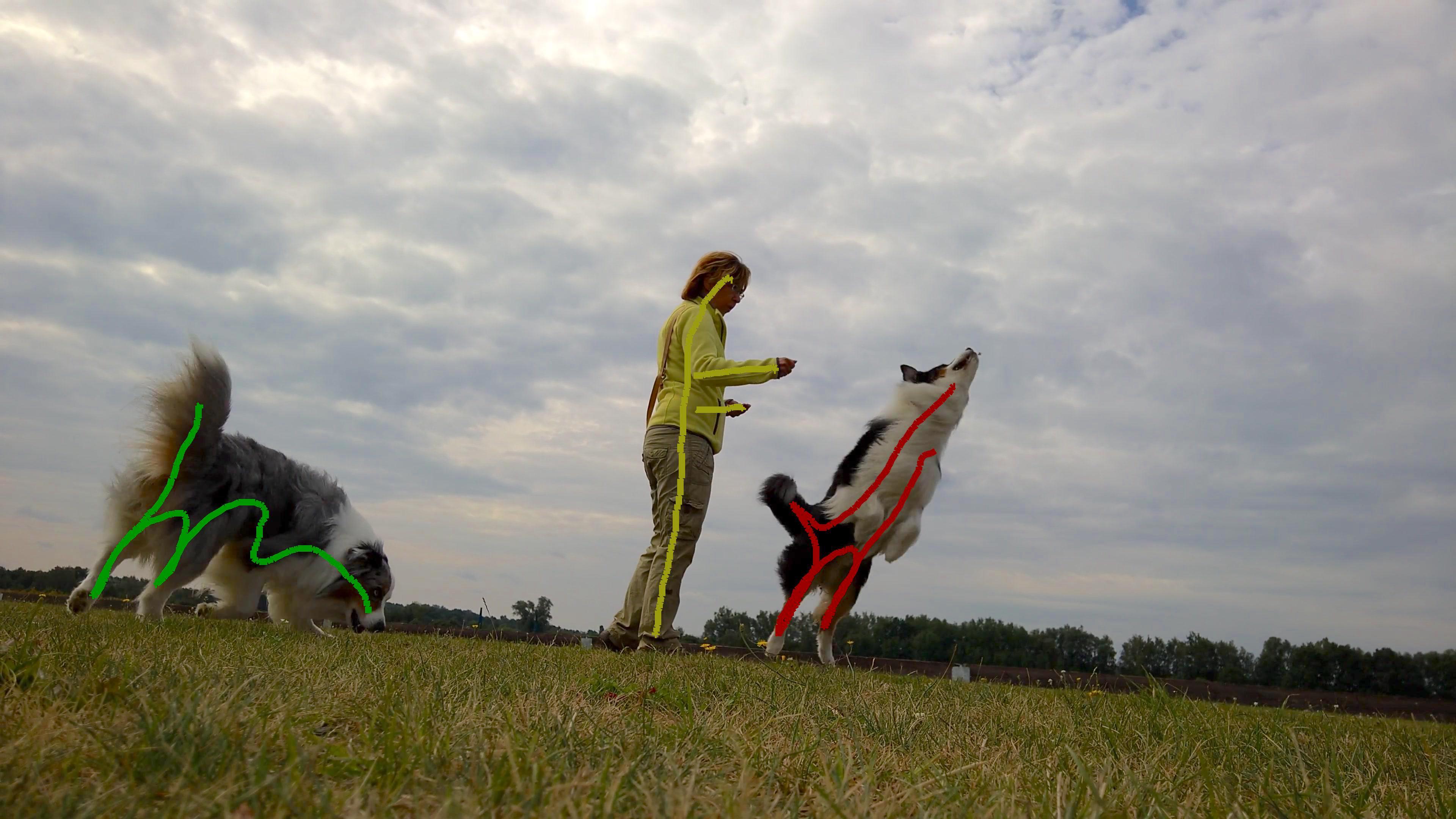}}};
    \draw (-1, 1.1) node[color=blue] {8.1 seconds};
\end{tikzpicture}\hfill
\begin{tikzpicture}
    \draw (0, 0) node[inner sep=0] {\fbox{\adjincludegraphics[width=0.49\linewidth,trim={{.03\width} {.1\height} {.3\width} {.1\height}},clip]{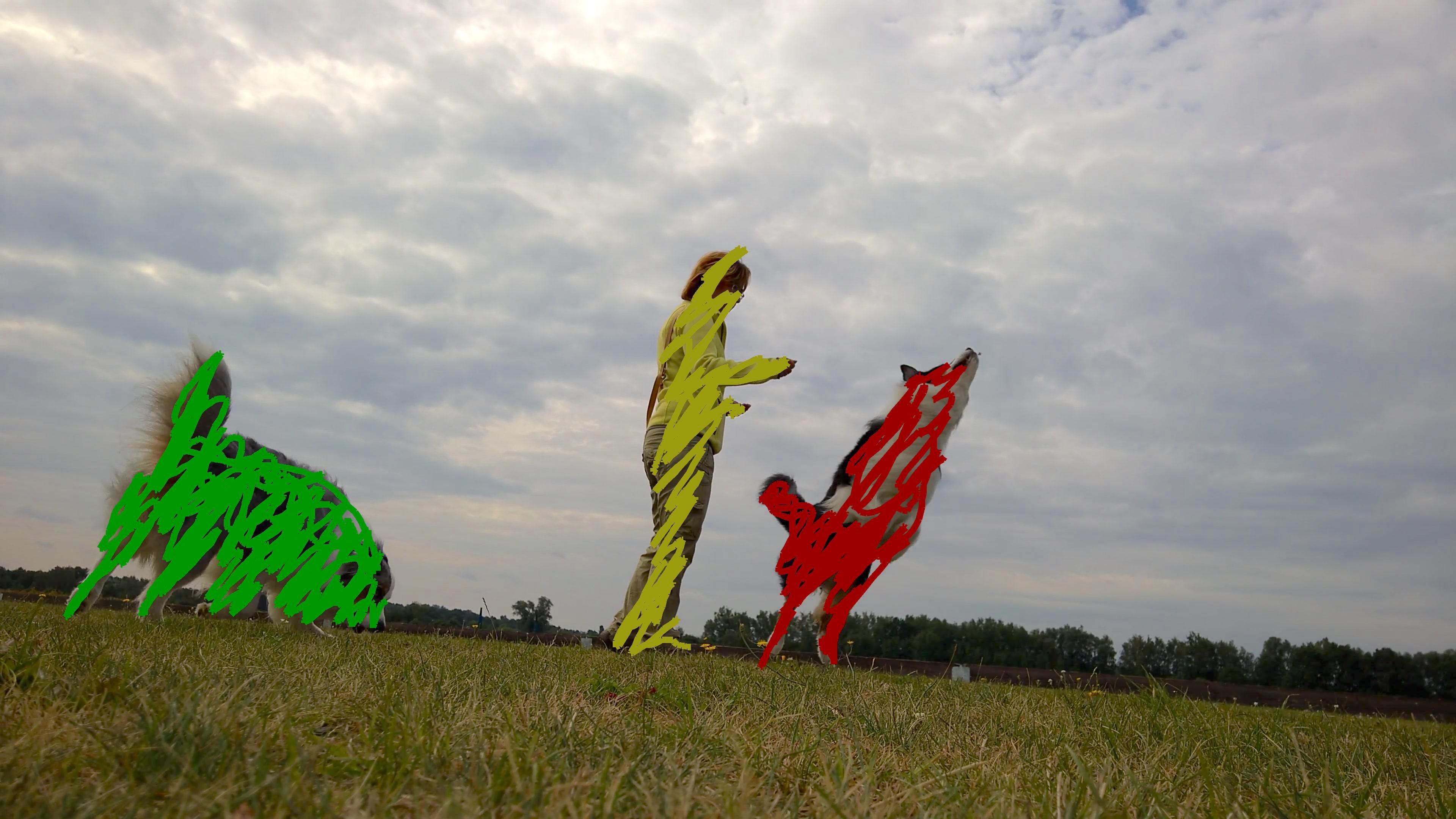}}};
    \draw (-1, 1.1) node[color=blue] {30.8 seconds};
\end{tikzpicture}
\caption{\textbf{Different levels of interaction in video object segmentation:} Top left unsupervised, top right semi-supervised; bottom interactive segmentation with different levels of detail.}
\end{figure}

The evaluation then measures the evolution of the quality of the result with respect to \textit{wall} time, that is, taking into account how many interactions the user needs, as well as how much time the algorithm takes to respond to each iteration, in order to reach a certain result quality.
We believe this scenario is more realistic than the semi- and unsupervised setups, both for the more natural inputs provided, as well as for the fact that time is taken into account.
Figure~\ref{fig:levels_of_interaction} illustrates different kinds of input annotations that one can obtain, given different annotation costs.

Interactive segmentation, however, poses some evaluation challenges. First, the results depend on the set of interactions provided by a human, so it is not easily scalable and reproducible; and the interactions depend on the set of results obtained, so one cannot rely on a set of pre-computed interactions. To solve this first challenge, we propose an automatic algorithm that simulates human interaction, providing a set of inputs to correct a certain result.
Second, in order to keep the test-set annotations of the DAVIS Challenge private, we present a web service to which a local process can connect, from which some inputs are received, and to which the segmentation results are sent. During a segmentation \textit{session} the service monitors the evolution of the quality of the results, as well as the time that the algorithm took to compute them.

To demonstrate the usefulness of our framework, we propose some powerful interactive segmentation baselines and a set of metrics to evaluate the results.
We believe that this framework can further promote research in the field of video object segmentation and help both research laboratories and industry discover better algorithms that can be used in realistic scenarios.

\section{Semi-Supervised Video Object Segmentation}
The main track of the 2018 edition of the DAVIS Challenge will be, as in the 2017 edition, the semi-supervised scenario, that is, the setup in which the perfect segmentation of the objects in the first frame is given to the algorithms, and the segmentation in the rest of the frames is the expected output.
We will also keep the same dataset and splits: training (60 sequences), validation (30 sequences), test-dev (30 sequences), and test-challenge (30 sequences).
Both the images and the full annotations are public for the training and validation sets.
For both test sets, only the annotations in the first frame are publicly available. 
The evaluation server for test-dev is always open and accepts unlimited submissions, whereas the submissions to test-challenge are limited in number (5) and time (2 weeks).

The detailed evaluation metrics are available in the 2017 edition document~\cite{Pont-Tuset_arXiv_2017}, and detailed dates and instructions can be obtained in the website of the challenge (\url{http://davischallenge.org/challenge2018/}).

\section{Interactive Video Object Segmentation}

\paragraph*{\textbf{Motivation}}
The current DAVIS benchmarks provide evaluation for the two extreme kinds of labels: pixel-wise segmented masks, and no labels at all. However, none of these scenarios are realistic in practice: detailed masks are tedious to acquire (79 seconds per instance on the coarse polygon annotations of COCO~\cite{Lin+14}, significantly more for DAVIS-level quality) and  unsupervised methods have no guiding signal for the user to select the object of interest, which is especially problematic in the multiple-object case.

In order to overcome these limitations, and make video object segmentation applicable in practice, we focus on a middle-ground solution: interactive segmentation using scribble supervision.
In this scenario, the user is given a raw video in the form of a canvas of images. Their task is to gradually refine the output of a method, interactively, by drawing scribbles on the object that needs to be segmented.

Different than the semi-supervised case, the user has access to the current results of their method, and the goal is to refine them. Moreover, the labelling is not limited to the first frame. For example, if a method fails at segmenting a particular object in frame $n$, the user can draw an additional scribble on that frame, and provide it to the method for additional processing. We show that using such interactive process builds up to the performance of the fully supervised case, in a much shorter labelling time.

\begin{figure*}
\centering
\adjincludegraphics[width=1\linewidth]{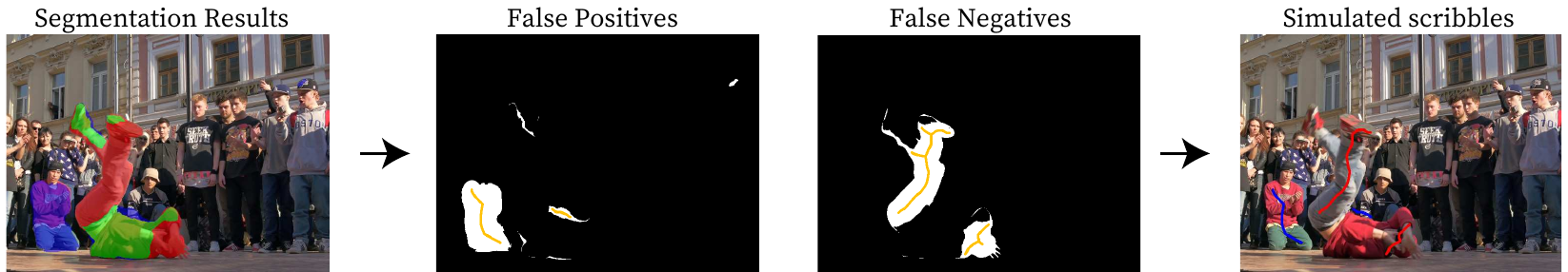}
\caption{\textbf{Simulated scribbles}: First the frames are evaluated obtaining true positives (green), false negatives (red) and false positives (blue). Then, for each class, the skeleton tree is computed from error regions (yellow). The final scribbles are obtained using the largest path in the skeleton tree.}
\label{fig:robot}
\end{figure*}

\paragraph*{\textbf{Evaluation as a Web Service}}

Having the initial prediction of a method, humans would evaluate all frames and provide extra scribbles in the region(s) where the prediction is poor. 
However, using humans in the loop is not feasible for large-scale experiments, so we simulate the human annotations by a robot.
We propose a fully automatic web service that evaluates segmentation techniques by directly interacting with the client computer, which allows the DAVIS organizers to keep the test annotations private.

More specifically, the interactive segmentation pipeline starts on the client side, by the user, who contacts the server and requests a video with one or multiple objects, and their initial scribbles. Since the first scribbles are independent of any prediction outcome, the server provides real ones taken from a pool of crowd-sourced initial scribbles. Once these are received, the user can run their method and return an initial prediction to the server. The task of the server is then to evaluate the prediction, and generate additional simulated scribbles. This way, we are able to validate interactive video object segmentation in large-scale datasets, the only task of the user being to interface their segmentation method to communicate with the server.

\paragraph*{\textbf{Crowd-sourced initial scribbles}}
We conduct a small-scale crowd-sourcing experiment, in which the annotators are first asked to label the objects of DAVIS 2017 with scribbles, on the frame that best represents them.
The collected scribbles can be used as the first interaction in the segmentation pipeline.
Figure~\ref{fig:levels_of_interaction} (bottom row) shows examples of these manually drawn scribbles.
Users are also asked to perform some iterations of the interactive video object segmentation baselines
presented below.
We also keep track of the time that people spend to draw the scribbles, for estimating the labelling cost.

\paragraph*{\textbf{Scribble simulation}}
After the first scribble has been drawn, the segmentation algorithm being evaluated obtains a first set of predictions on the full video.

The server, which has access to the ground-truth masks, focuses on the frame where the evaluation metric between the prediction and the ground truth is the worst, and provides a new scribble in that frame, for the next iteration.

The region for which the next scribble needs to be generated is defined by simple binary operations, by first removing small spurious detections and selecting connected components where the prediction has gone wrong.
We define the generated scribbles as a simplified skeleton of such regions.
The process is repeated for both false positives and false negatives.
Figure~\ref{fig:robot} illustrates a qualitative example of a simulated scribble set.
One can easily calculate the labelling costs of the scribbles by estimating it from the crowd-sourced ones.

\paragraph*{\textbf{Evaluation metrics}}
The key aspect we want to evaluate in interactive segmentation is the trade off between interaction time and accuracy of the result. 
The interaction time is composed of the time the user needs to provide the feedback and the time the algorithm takes to respond with a new segmentation result.
The former will be estimated by the scribbling robot, having as reference the times of the collected user scribbles, and the latter will be measured by the evaluation service as the time the client process takes to submit the new segmentation result.

As accuracy metric we use the $\mathcal{J}\&\mathcal{F}$, as presented and motivated in the 2017 DAVIS challenge~\cite{Pont-Tuset_arXiv_2017}.
Each run of interaction will therefore be represented by a sequence of time stamps and accuracy values.
To aggregate this sequence into a final quality metric to compare the participants' performance, we propose two different tracks in the competition:
the quality track, in which we are willing to wait a reasonable amount of time to obtain an accurate result; and the speed track, in which we want a result as soon as possible but we expect a minimum accuracy in each object.

In the quality track, we set a time budget, for example $5$ seconds per frame per object, and we compute the $\mathcal{J}\&\mathcal{F}$ that a method can reach when spending the whole time budget.
%For example in a sequence with $2$ objects and $50$ frames, the total amount of time that a method can spend taking into account all the interactions is $500$ seconds. In order to mimic a realistic scenario, the time budget is fixed for every sequence, but the final accuracy is computed as the mean accuracy of all the objects in a specific set \ie test-dev to give the same relevance to small and big objects. For example as shown in~\ref{fig:qual_vs_time}, OSVOS achieves $61.5\% \mathcal{J}\&\mathcal{F}$.
In the fast track, we set a minimum quality, for instance $60\%$ in $\mathcal{J}\&\mathcal{F}$, and we want to know how much time an algorithm needs to achieve at least that quality in each of the objects.
%If for a certain object in a sequence the method never reaches the minimum quality, we would compute the time spent on that object in that sequence as $5$ seconds per frame.
The final summary measure is the sum of the time spent in each of the objects.

We believe that these two evaluation metrics will encourage the video object segmentation community to not only focus on the accuracy of their methods, but also in their speed, in order to make methods more usable in realistic scenarios.  

\paragraph*{\textbf{Baseline methods}}
We define two baseline methods to show the usefulness of the proposed benchmark and metrics and to serve to the participants to the challenge as a reference and guide to develop their own techniques.
We will make the code of these baselines available so that they can serve as a starting point.

The first baseline is based on the OSVOS~\cite{Caelles2017} technique for semi-supervised segmentation. OSVOS updates the appearance model of the target object by fine-tuning a CNN on the first mask of the object, and processes each frame of the video independently, making the technique efficient in terms of speed. 

In our scenario, OSVOS is adapted to have cheaper annotations in the form of scribbles as its input, instead of the full pixel-wise labelled mask. We create a weaker version of the ground truth from the scribbles. Specifically, if $X$ is the scribble, we assign the foreground label to the pixels of $X_{fg} = X \oplus B$, where $B$ is a structuring element and $\oplus$ the dilation operator. The pixels of $X_{nc} = X \oplus C \setminus X_{fg}$, with $C \supseteq B$, are assigned a \textit{no-care} label that is excluded from the loss when fine-tuning, and the rest of the pixels are treated as background. In their original paper, the authors validated OSVOS on the DAVIS 2016 dataset of single-object sequences. Since we work on DAVIS 2017 with multiple objects, we experiment on a single object per fine-tuning step. We treat scribbles on the other objects as background, when available.
We refer to the modified version of OSVOS as Scribble-OSVOS.

Typical video segmentation methods report per-frame results, dividing the total amount of processing time by the number of frames. In the interactive case, however, in order to proceed, we need results for the entire sequence to be available. A fine-tuning step of OSVOS can vary from 60 seconds to 15 minutes~\cite{Caelles2017}, depending on the quality we want to achieve. We examine the fastest version of the algorithm (60 seconds), to keep waiting time between the interactions as low as possible.

We start by manually drawing one scribble per object, from which an initial prediction is generated, by running Scribble-OSVOS. Afterwards, given the prediction, we draw additional foreground and background scribbles in erroneous areas, and re-train Scribble-OSVOS. This process is repeated until we reach a satisfactory result. Note that current foreground predictions are marked as \textit{no-care} areas, in order to avoid training with noisy labels. 

The second baseline is specifically designed to avoid the expensive retraining step every time the user introduces a new scribble, which adds a significant lag. Even though OSVOS results in satisfactory qualitative results, the user has to wait for at least 60 seconds to evaluate the results of their previous labelling step. When one has to repeat this process for many sequences, this process becomes tedious and out of the scope of interactive segmentation.

To improve user experience and reduce the lag introduced between the interactions, the idea is to use a CNN to obtain a feature vector for each pixel of the video and to train a linear classifier from the pixels that are annotated using the scribbles. In our baseline, we use Deeplab-v2~\cite{Che+17} with the \textit{ResNet-101}~\cite{He+16} architecture, which has been trained for semantic segmentation. The output size of the CNN for every frame is the size of the input downsampled by a factor of 8 and each pixel has a 2048-feature vector. 
We draw inspiration from~\cite{Chen2018}, which further trains the CNN embedding by means of pixel-wise metric learning.

We handle the scribbles in the same way than in the Scribble-OSVOS baseline, with the same pre-processing. 
Using these labels and the feature vectors from the annotated frames, we train a support vector classifier~\cite{Fan+08} for every object.

At test time, we also extract the feature vector for every pixel, without need to recompute it on each interaction.
For every object, we use its classifier to label every pixel as foreground or background.
Every time a new scribble is drawn, we only need to adapt the support vector classifier.
Note that Scribble-OSVOS, in contrast, adapts the features and the classifier (last layers of the CNN) on each interaction.

\begin{figure}[h]
\centering
\resizebox{\linewidth}{!}{\begin{tikzpicture}[/pgfplots/width=1.01\linewidth, /pgfplots/height=0.7\linewidth, /pgfplots/legend pos=south east]
    \begin{axis}[ymin=35,ymax=65,xmin=0,xmax=450,enlargelimits=false,
        xlabel=Total timing (seconds),
        ylabel= $\mathcal{J}\&\mathcal{F}(\%)$,
		font=\scriptsize,
        grid=both,
		grid style=dotted,
        xlabel shift={-2pt},
        ylabel shift={-5pt},
        %xmode=log,
        legend columns=1,
        %transpose legend,
        legend style={/tikz/every even column/.append style={column sep=3mm}},
        minor ytick={2,4,...,70},
        minor xtick={0,20,...,450},
        ytick={10,20,...,90},
        legend pos= south east
        ]

        \addplot+[smooth, Set2-8-1, mark=none, line width=1.5] table[x=seconds,y=J&F] {data/interactive/scrib-osvos.txt};
        \addlegendentry{Scribble-OSVOS}
        \label{fig:qual_vs_time:scribble-osvos}
        
        \addplot+[smooth, Set2-8-2, mark=none, line width=1] table[x=seconds,y=J&F] {data/interactive/scrib-robot-osvos.txt};
        \addlegendentry{Simulated-Scribble-OSVOS}
        \label{fig:qual_vs_time:scribble-robot-osvos}

        \addplot+[smooth, Set2-8-4, mark=none, line width=1.5] table[x=seconds,y=J&F] {data/interactive/scrib-linear.txt};
        \addlegendentry{Scribble-Linear}
        \label{fig:qual_vs_time:scribble-linear}
        
        \addplot+[smooth, Set2-8-6, mark=none, line width=1] table[x=seconds,y=J&F] {data/interactive/scrib-robot-linear.txt};
        \addlegendentry{Simulated-Scribble-Linear}
        \label{fig:qual_vs_time:scribble-robot-linear}
        
%        \addplot+[green,mark=none, line width=1] table[x=seconds,y=J&F] {data/interactive/scrib-linear-bilateral-optimized.txt};
%        \addlegendentry{Linear-bilateral-opt}
        
%        \addplot+[black,mark=none, line width=1] table[x=seconds,y=J&F] {data/interactive/scrib-linear-bilateral-plain.txt};
%        \addlegendentry{Linear-bilateral}
%        
%        	\addplot+[cyan,fill=cyan,mark=*, mark size=1.5,only marks, mark options={fill=cyan}] coordinates{(1200, 57)};
%        \addlegendentry{OSVOS}
%        
%        	\addplot+[magenta,fill=cyan,mark=*, mark size=1.5,only marks, mark options={fill=magenta}] coordinates{(120, 39.26)};
%        \addlegendentry{Linear}

    \end{axis}
\end{tikzpicture}}
   \caption{\textbf{Quality vs. Timing:} Evolution of $\J \& \F$ in DAVIS 2017 validation set as a function of the available time.}
   \label{fig:qual_vs_time}
\end{figure}
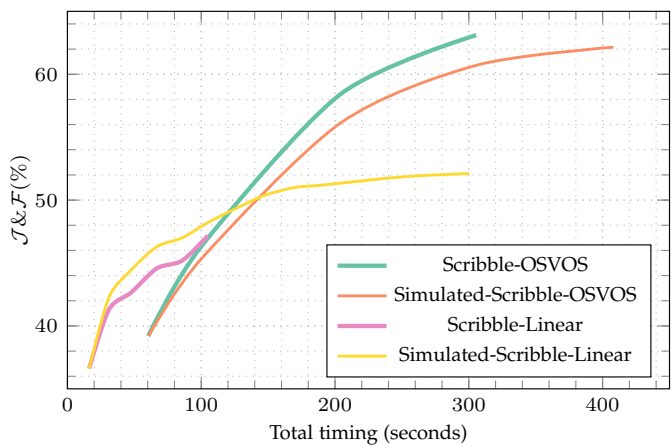

\paragraph*{\textbf{Evaluation results}}
Figure~\ref{fig:qual_vs_time} illustrates the comparison of manually annotated scribbles to simulated ones, for both aforementioned baselines.

Scribble-OSVOS when trained from manual scribbles~(\ref{fig:qual_vs_time:scribble-osvos}) builds performance marginally faster than its simulated counterpart~(\ref{fig:qual_vs_time:scribble-robot-osvos}). The support vector classifier follows the opposite trend (\ref{fig:qual_vs_time:scribble-linear} vs. \ref{fig:qual_vs_time:scribble-robot-linear}). Results indicate that simulated scribbles allow us to extract similar conclusions as the ones generated by humans, in both baselines, which we believe justifies the use of our simulated scribble generator in the interactive video object segmentation challenge.

Interestingly, performance of OSVOS, with a pixel-wise annotated label (full supervision) achieves 57\% in 1200 seconds, a number that is surpassed by Scribble-OSVOS in only 200 seconds of wall time.
The same happens for the support vector classifier, where the scribble-supervised algorithm reaches the performance of the fully supervised one (39.26\% in 110 seconds) in only 13 seconds.

\section{Conclusions}
This paper present the 2018 DAVIS Challenge on Video Object Segmentation. The main track will be the same format and same dataset as in the previous edition: semi-supervised segmentation by taking the perfect mask of the first frame as input.

We also propose a new teaser track: interactive video segmentation.
In order to simulate the user interaction while keeping the test annotations private, we propose a web service to which the participants will connect and from which they will be able to obtain a sequence of scribbles to which they will reply by submitting the resulting segmentation.
By measuring the trade off between response speed and quality of the results we aim to encourage research into the field of interactive video object segmentation, which we believe is the cornerstone to making video object segmentation usable in realistic scenarios.

% use section* for acknowledgment
\ifCLASSOPTIONcompsoc
  % The Computer Society usually uses the plural form
  \section*{Acknowledgments}
\else
  % regular IEEE prefers the singular form
  \section*{Acknowledgment}
\fi

Research partially funded by the workshop sponsors: Google, Disney Research, Prof. Luc Van Gool's Computer Vision Lab at ETHZ, and Prof. Fuxin Li's group at the Oregon State University. The authors gratefully acknowledge support by armasuisse, and thank NVIDIA Corporation for donating the GPUs used in this project.

% Can use something like this to put references on a page
% by themselves when using endfloat and the captionsoff option.
\ifCLASSOPTIONcaptionsoff
  \newpage
\fi

\bibliographystyle{IEEEtran}
% argument is your BibTeX string definitions and bibliography database(s)

\bibliography{DAVIS2018}

\end{document}